\documentclass[final,5p,times,twocolumn]{elsarticle}

\usepackage{amssymb}
\usepackage{amsmath}

\usepackage{adjustbox}
\usepackage{graphicx}
\usepackage{booktabs}
\usepackage{color}
\usepackage{colortbl}
\usepackage{hyperref}
\usepackage{url}

\usepackage{amsfonts}
\usepackage{tabularx}
\usepackage{booktabs} 
\usepackage{pifont}
\usepackage{bbm}
\usepackage{tcolorbox}
\usepackage[inline]{enumitem}

\usepackage{xspace}

\newcommand{\ours}{WISH\xspace}
\definecolor{mygreen}{rgb}{0.533, 0.906, 0.533}
\definecolor{myred}{rgb}{0.973, 0.514, 0.475}
\definecolor{mylightgray}{rgb}{0.9, 0.9, 0.9}

\journal{Pattern Recognition Letters}

\begin{document}

\begin{frontmatter}



\title{Learning Egocentric In-Hand Object Segmentation through Weak Supervision from Human Narrations}


\author[isti]{Nicola Messina} 
\ead{nicola.messina@isti.cnr.it}
\author[unict]{Rosario Leonardi}
\ead{rosario.leonardi@unict.it}
\author[isti]{Luca Ciampi}
\ead{luca.ciampi@isti.cnr.it}
\author[isti]{Fabio Carrara}
\ead{fabio.carrara@isti.cnr.it}
\author[unict]{Giovanni Maria Farinella}
\ead{giovanni.farinella@unict.it}
\author[isti]{Fabrizio Falchi}
\ead{fabrizio.falchi@isti.cnr.it}
\author[unict]{Antonino Furnari}
\ead{antonino.furnari@unict.it}

\affiliation[isti]{organization={Institute of Information Science and Technologies - National Research Council},
            city={Pisa},
            country={Italy}}

\affiliation[unict]{organization={University of Catania},
            city={Catania},
            country={Italy}}

\begin{abstract}
Pixel-level recognition of objects manipulated by the user from egocentric images enables key applications spanning assistive technologies, industrial safety, and activity monitoring. 
However, progress in this area is currently hindered by the scarcity of annotated datasets, as existing approaches rely on costly manual labels. 
In this paper, we propose to learn human-object interaction detection leveraging \textit{narrations} -- natural language descriptions of the actions performed by the camera wearer which contain clues about manipulated objects. 
We introduce \textit{\textbf{N}arration-\textbf{S}upervised \textbf{i}n-\textbf{H}and \textbf{O}bject \textbf{S}egmentation} (NS-iHOS), a novel task where models have to learn to segment in-hand objects by learning from natural-language narrations in a weakly-supervised regime. Narrations are then not employed at inference time.
We showcase the potential of the task by proposing \textit{\textbf{W}eakly-Supervised \textbf{I}n-hand Object \textbf{S}egmentation from \textbf{H}uman Narrations} (\ours), an end-to-end model distilling knowledge from narrations to learn plausible hand-object associations and enable in-hand object segmentation without using narrations at test time. 
We benchmark \ours against different baselines based on open-vocabulary object detectors and vision-language models. Experiments on EPIC-Kitchens and Ego4D show that \ours surpasses all baselines, recovering more than $50\%$ of the performance of fully supervised methods, without employing fine-grained pixel-wise annotations. Code and data can be found at \href{https://fpv-iplab.github.io/WISH/}{\url{https://fpv-iplab.github.io/WISH/}}.
\end{abstract}




\end{frontmatter}



\section{Introduction}
\label{sec:intro}

Egocentric vision offers a privileged view into human-object interactions, enabling a fine-grained understanding of which objects are manipulated and how the user engages with them from their first-person perspective.
This ability to detect and segment in-hand objects, those actively manipulated by the user, can enable intelligent systems to gain a deeper understanding of human behavior, offering valuable insights into how individuals engage with their surroundings and supporting a wide range of applications~\cite{Plizzari2024AnOutlook}, from daily activity analysis~\cite{VISOR2022} to assistive technologies~\cite{betancourt2015evolution}, robot-assisted surgeries~\cite{DBLP:conf/miccai/WangKZML23}, and industrial safety~\cite{ragusa2024enigma}.
\begin{figure}[t]
     \centering
     \includegraphics[width=\linewidth]{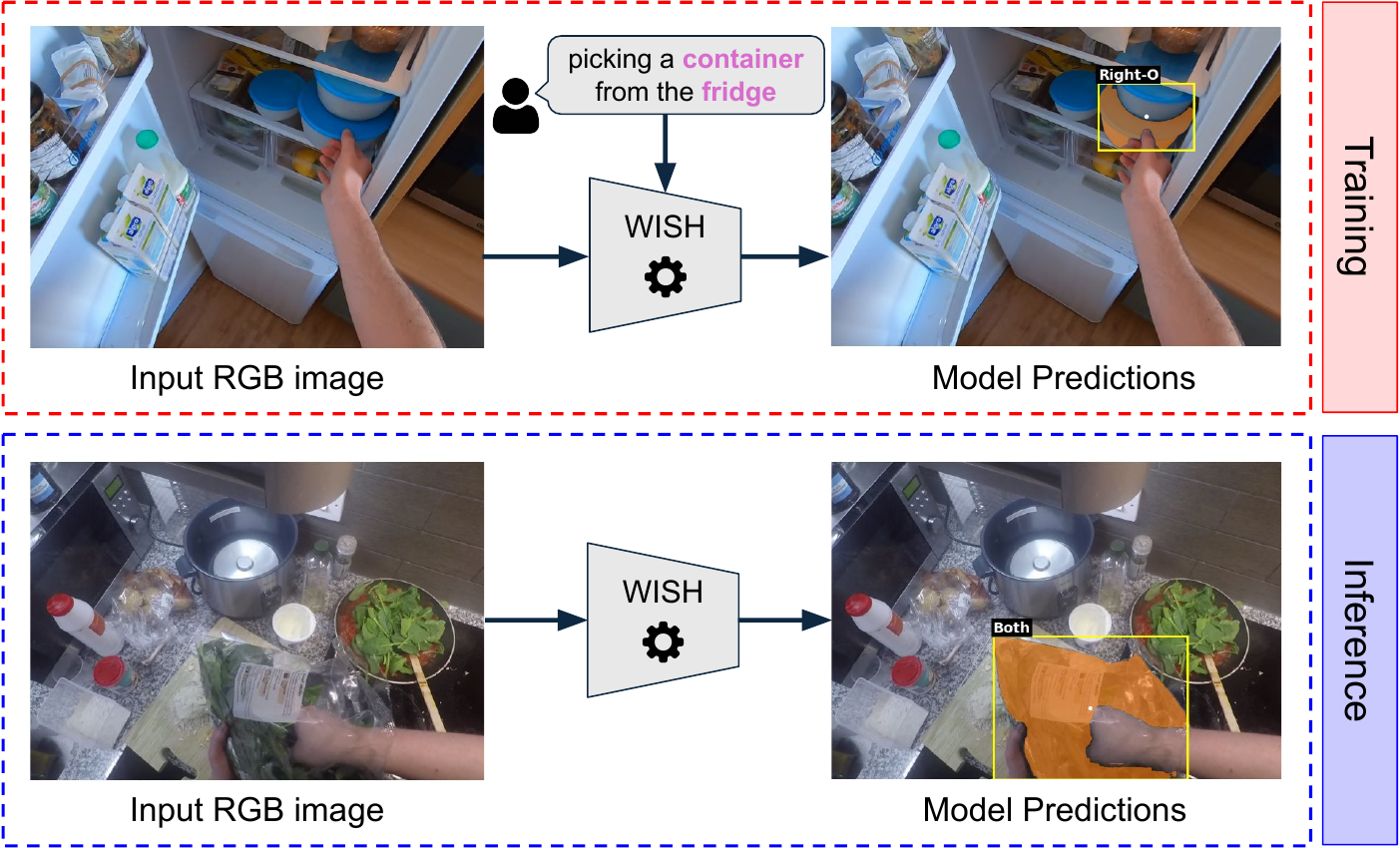}
     \caption{Models designed to solve the NS-iHOS task, like our proposed \ours, learn to segment in-hand objects using human narrations as the sole form of weak supervision; predictions are then made directly from images at test time.}
     \label{fig:teaser_wish}
 \end{figure}
Despite the widespread availability of wearable devices, detecting manipulated objects remains significantly underexplored. This is largely due to the scarcity of public datasets and the high cost associated with acquiring high-quality annotations. Most existing approaches typically rely on fully supervised learning~\cite{Shan2020UnderstandingHH, VISOR2022, EgoHos_jianbo_eccv22,ragusa2024enigma}, requiring detailed manual annotations. 
Advancing research in this area needs novel strategies that reduce the reliance on manual labeling and facilitates scalable data collection in real-world scenarios.

In this paper, we explore a different direction to tackle the lack of suitable fully annotated datasets for detecting in-hand objects. 
In particular, the most prominent large-scale datasets of egocentric videos~\cite{Damen2021RESCALING,Grauman2021Ego4DAT} include \textit{narrations} --- densely-annotated, temporally-localized natural language descriptions of the events within each video moment, which are cheaper to obtain at scale compared to pixel-wise labels.
Crucially, as descriptions of the user's interactions with the environment, narrations inherently mention the specific objects the camera wearer interacts with, offering a reliable source of weak supervision that scales effectively with the size of modern datasets.
Towards this direction, we introduce \textit{\textbf{N}arration-\textbf{S}upervised \textbf{i}n-\textbf{H}and \textbf{O}bject \textbf{S}egmentation} (NS-iHOS), a novel task where models learn to recognize and segment objects manipulated by the camera wearer from egocentric images using paired narrations as supervision at training time, but requiring only image inputs at test time. 
NS-iHOS builds on previous formulations~\cite{banerjee2024hot3d,EgoHos_jianbo_eccv22}, where methods were tasked with segmenting in-hand objects given full supervision. Unlike these works, we propose scaling these methodologies beyond the bottleneck of carefully crafted manual labels.

To tackle the proposed NS-iHOS task, we propose 
\ours (\textit{\textbf{W}eakly Supervised \textbf{I}n-Hand Object \textbf{S}egmentation from \textbf{H}uman Narrations}), an architecture designed to combine the supervisory signal of nouns mentioned in the narrations with the weak spatial understanding of CLIP~\cite{radford2021learning} to determine whether hands are interacting with any of the objects present in the scene. \ours integrates two stages: 1) noun-object alignment, where a CLIP-based shared vision-language space matching nouns extracted from narrations and objects detected with an open-vocabulary detector is learned, and 2) hand-object matching, which distills pseudo-labels obtained from aligned noun-object pairs into a module capable of matching hands and objects, enabling the identification of in-hand objects at test time on raw images without relying anymore on narrations (see Figure~\ref{fig:wish_arch}).
We benchmark \ours against baselines combining open-vocabulary object detectors and Vision-Language Models (VLM), which offer a current understanding of what state-of-the-art multimodal language-vision approaches can achieve in this context. To mitigate the computational demand of large language models, we also investigate the possibility of using VLMs to process narrations and generate pseudo-labels, which can then be used to train a more compact, off-the-shelf HOI detection models. Experiments on EPIC-Kitchens~\cite{Damen2021RESCALING,VISOR2022} and Ego4D~\cite{Grauman2021Ego4DAT,EgoHos_jianbo_eccv22} show that \ours outperforms all baselines and recovers more than $50\%$ of the performance of a fully supervised method without any access to pixel-wise segmentation masks. 

\section{Related Works}
\paragraph{Egocentric Human-Object Interaction Detection}
Human-Object Interaction detection has been extensively studied within the domain of third-person vision~\cite{chao:wacv2018}, where algorithms are tasked with detecting humans and their interactions~\cite{GkioxariDRHOI}.
Significant progress in this area has been driven by the availability of large-scale annotated datasets~\cite{everingham2009pascal,chao:wacv2018} and standardized evaluation protocols, enabling benchmarking and methodological advancements. In contrast, Egocentric Human-Object Interaction detection remains comparatively underexplored. 
In this context, the problem of \textit{Hand-Object Interaction} (HOI) detection was first formulated in~\cite{Shan2020UnderstandingHH} as a combined task involving the detection of hands their manipulated objects via bounding boxes, along with the estimation of hand contact states, and adopted by several follow-up studies~\cite{ragusa2024enigma, leonardi2022egocentric, leonardi2024exploiting}. The authors of~\cite{VISOR2022} proposed \textit{Hand-Object Segmentation} (HOS) as an alternative formulation of HOI detection, extending the detection objective to pixel-wise segmentation. 
Our NS-iHOS task builds on the formulation introduced in~\cite{banerjee2024hot3d}, which segments in-hand objects,  distinguishing whether they are held by the left hand, right hand, or both.
Only a few approaches aimed to reduce the reliance on fully-supervised data, mainly focusing on synthetic data generation~\cite{leonardi2025synthetic, ehsani2021manipulathor}.
In contrast to these works, we seek to reduce reliance on fully supervised annotations by exploiting weakly supervised learning with natural language narrations, enabling scalable training without extensive manual labeling.

\paragraph{Language-Guided Object Segmentation}
Contrastively-learned vision-language representations like CLIP \cite{radford2021learning} paved the way towards a deep interaction between images and text and enabled tasks like open-vocabulary image classification 
\cite{allgeuer2025unconstrained,zhai2023sigmoid} and image segmentation~\cite{ding2022decoupling,xu2023open}, where classes are treated as natural language text rather than discrete categorical items.
Similarly, other research branches focused on the weakly-supervised image-phrase grounding task \cite{chen2022contrastive,wang2020maf}, which requires localizing corresponding objects in an image given a phrase, or on fine-grained understanding of text-region correspondences from coarse image-caption pairs 
\cite{messina2021fine,pan2023fine}. 
Object grounding from natural language descriptions has also been recently investigated in egocentric vision~\cite{shen2024learning,bansal2024hoi}.
While the aforementioned works provide an important foundation, the NS-iHOS task requires models to correctly identify and segment the in-hand object, rather than a specific object category provided in natural language. This requires correctly associating the objects with the person's hands, and assumes the availability of narrations only at training time.

\section{NS-iHOS Task Definition and Benchmark} 
We frame the \textbf{N}arration-\textbf{S}upervised \textbf{i}n-\textbf{H}and \textbf{O}bject \textbf{S}egmentation (NS-iHOS) task as learning to segment objects manipulated by a user in egocentric images, relying solely on natural language narrations as supervision during the training phase.
This formulation is based on the premise that narrations describing a user’s actions inherently encode informative cues about the objects being handled. 
For example, a narration such as \textit{``picking a container from the fridge"} offers a high-level, yet indirect signal linking the visual scene to interacted objects. The challenge for the model is to translate this coarse semantic guidance into accurate, pixel-level segmentation masks.
At evaluation time, the model is required to perform inference using only the visual input, without access to the narration. 


\paragraph{Task Definition}
Let training and test sets be denoted as
\begin{equation}
\mathcal{D}_{\mathrm{train}} = \{(\mathbf{I}_i, \mathcal{N}_i)\}_{i=1}^{N_{\mathrm{train}}}, \quad
\mathcal{D}_{\mathrm{test}} = \{(\mathbf{I}_j, \{M_{e,j}^{\ast}\}_{e \in \mathcal{E}})\}_{j=1}^{N_{\mathrm{test}}},
\end{equation}
where $\mathbf{I}_i \in \mathbb{R}^{W \times H \times 3}$ is an egocentric image, $\mathcal{N}_i$ is its corresponding natural language narration, and $M_{e,j}^{\ast}$ denotes the ground-truth pixel-level mask for interaction type $e$ (available only for evaluation).
Let $\mathcal{E} = \{\mathrm{L}, \mathrm{R}, \mathrm{B}\}$ denote the interaction types: left hand, right hand, and both hands.
The goal of NS-iHOS is to learn a model parametrized by $\theta$: $
f_{\theta} : \mathbb{R}^{W \times H \times 3} \longrightarrow \{0,1\}^{W \times H \times |\mathcal{E}|}$.
At inference, for each $\mathbf{I} \in \mathcal{D}_{\mathrm{test}}$, we obtain
\begin{equation}
\{ M_{e} \}_{e \in \mathcal{E}} = f_{\theta}(\mathbf{I}), \quad  M_{e} \in \{0,1\}^{W \times H},
\end{equation}
where $M_{\mathrm{L}}$, $M_{\mathrm{R}}$, and $M_{\mathrm{B}}$ are the predicted binary masks for objects interacted with the left hand, right hand, and both hands, respectively, to be compared with ground truth $M^*_L, M^*_R, M^*_B$ for evaluation.
The narration $\mathcal{N}_i$ is used only as supervision during training, whereas at inference, $f_{\theta}$ depends solely on the image.

\paragraph{Benchmark}
To evaluate the proposed NS-iHOS task, we propose a benchmark based on two large-scale egocentric datasets: \textit{EPIC-Kitchens}~\cite{damen2018scaling, Damen2021RESCALING} and \textit{Ego4D}~\cite{Grauman2021Ego4DAT}. For \textit{EPIC-Kitchens}, we adopt the official \textit{EPIC-Kitchens VISOR}~\cite{VISOR2022} train/val splits. 
Specifically, we source narrations from EPIC-Kitchens and ground truth pixel-level annotations of hand and active objects from VISOR. This set contains kitchen-related scenes and interactions with different food items and kitchen tools.
For \textit{Ego4D}, we construct our training set using the official annotations from the \textit{Hands and Objects} benchmark, 
which provides images of hand-object interactions and narrations. 
For evaluation, we rely on the subset of \textit{EgoHOS}~\cite{EgoHos_jianbo_eccv22} in overlap with Ego4D.
From this set, we obtain ground truth pixel-wise masks for hands and interacted objects, while associated narrations are not provided.
We ensure that our training set is fully disjoint from these splits to maintain a strict separation between training and evaluation data.
This set includes images from different scenarios, inheriting the diversity of Ego4D.
The EPIC-Kitchens subset of our benchmark contains $32,857$ training and $7,747$ test image/narration pairs, whereas the Ego4D subset contains $26,512$ training and $1,499$ val/test images. 

\begin{figure*}[t]
\centering
\includegraphics[width=\linewidth]{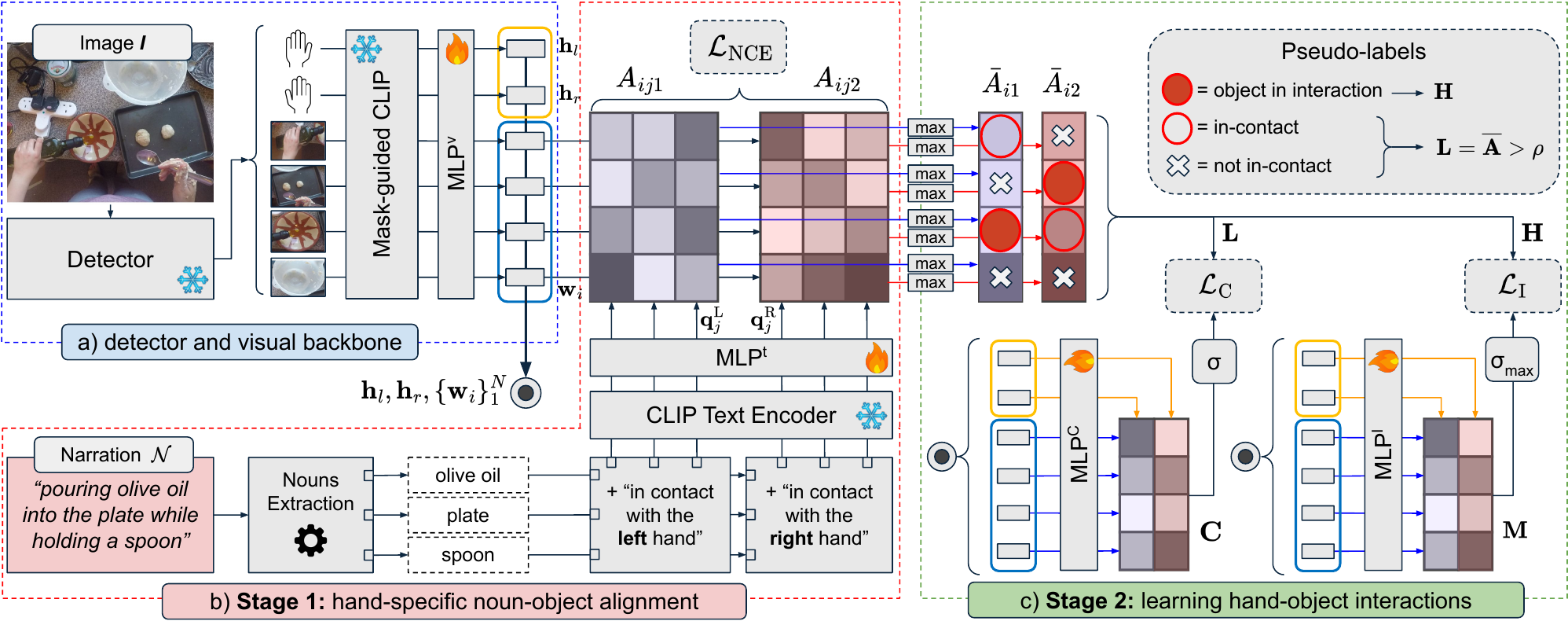}
\caption{\textbf{The Architecture of \ours} Our model operates in two stages sharing a common backbone. \textbf{(a)} An object segmenter and a CLIP-based backbone extract visual embeddings for all object and hand proposals. \textbf{(b)} In Stage 1, we learn a shared embedding space to align hand-specific noun phrases from narrations with their corresponding visual object embeddings. \textbf{(c)} In Stage 2, we generate pseudo-labels from this alignment to train two specialized heads: a Contactness head (C) and a Matching head (M). At test time, only the backbone and Stage 2 are used for narration-free in-hand object segmentation.}
\label{fig:wish_arch}
\end{figure*}

\paragraph{Evaluation Metrics}
We evaluate each model by comparing the predicted masks $(M_{\mathrm{L}}, M_{\mathrm{R}}, M_{\mathrm{B}})$ with the corresponding ground-truth masks $(M^{\ast}_{\mathrm{L}}, M^{\ast}_{\mathrm{R}}, M^{\ast}_{\mathrm{B}})$. 
Following the evaluation protocol of~\cite{EgoHos_jianbo_eccv22,banerjee2024hot3d}, we report the \textit{mean Intersection-over-Union} (mIoU) computed over the three interaction classes $\mathcal{E} = \{\mathrm{L}, \mathrm{R}, \mathrm{B}\}$.
In addition to class-specific evaluation, we report performance on an auxiliary \textit{Either} (E) class, defined by matching ground-truth and predicted object masks irrespective of the interacting hand. This aggregated category enables assessment of the model’s ability to detect in-hand objects independently of its accuracy in assigning them to a specific hand.

\section{\ours} 
To tackle the proposed NS-iHOS task, we propose an end-to-end weakly-supervised model called \textbf{W}eakly-Supervised \textbf{I}n-hand Object \textbf{S}egmentation from \textbf{H}uman Narrations (\ours).
\ours is guided by two assumptions derived from common human-object interaction datasets~\cite{VISOR2022,EgoHos_jianbo_eccv22}: (i) narrations describing an activity typically mention the objects being manipulated, grounding the text in the visual scene; and (ii) a person's hand typically interacts with at most one object at any given moment.

Based on these priors, \ours employs a two-stage architecture (Figure~\ref{fig:wish_arch}) that first learns a rich vision-language alignment space and then distills this knowledge into a lightweight, vision-only inference model. Specifically:
\begin{enumerate*}[label=\roman*)]
    \item \textbf{Stage 1: Hand-specific Alignment.} This stage acts as a bridge between the textual and visual domains. It learns to associate specific noun phrases from the narration (e.g., ``spoon") with their corresponding visual object segments in the image, while also discerning which hand (left or right) is involved in the interaction.
    \item \textbf{Stage 2: Hand-Object Interaction Learning.} This stage learns to perform the final NS-iHOS task without relying on narrations. It uses the alignments learned in Stage 1 to generate pseudo-labels, effectively teaching itself to identify which objects are being held and by which hand, using only visual cues.
\end{enumerate*}
This two-stage design allows \ours to leverage textual supervision during training while performing efficient, narration-free segmentation at inference time. 

\subsection{Visual and Textual Representation}

Our architecture shares a visual backbone across both stages (Figure~\ref{fig:wish_arch}{a}). We extract visual embeddings for candidate objects and hands using a pre-trained backbone. Specifically, we employ a mask-guided variant of the CLIP image encoder~\cite{shen2024learning} which only attends segmented areas. To adapt these general-purpose features to our task, we project each object embedding $\mathbf{v}_i$ and hand embedding $\mathbf{g}_{k}$ through a learnable MLP adapter ($MLP^\mathbf{v}$), yielding the final visual representations $\mathbf{w}_i = MLP^v(\mathbf{v}_i)$ and $\mathbf{h}_k = MLP^v(\mathbf{g}_k)$.

To bridge the visual representations with the narration $\mathcal{N}$, we extract noun phrases $\{\mathcal{P}_{j}\}$ and augment them to explicitly encode hand information. Specifically, we generate hand-specific prompts using templates $\mathcal{F}^L$ ("[noun] in contact with the left hand") and $\mathcal{F}^R$ ("[noun] in contact with the right hand"). These augmented phrases are encoded by the CLIP text encoder and refined via a learnable textual adapter ($MLP^t$) to produce hand-specific textual embeddings $\mathbf{q}_j^{(L,R)}$. Further details are provided in the Appendix.

\subsection{Stage 1: Hand-Specific Alignment}

The primary goal of this stage (Figure~\ref{fig:wish_arch}{b}) is to learn the parameters of the visual and textual adapters, $\mathrm{MLP}^v$ and $\mathrm{MLP}^t$, so that embeddings of visually present objects align closely with the corresponding hand-specific noun phrases extracted from the narration. 
For example, if the user is holding a spoon in their right hand, the visual embedding $\mathbf{w}_i$ of that spoon should exhibit high cosine similarity with the textual embedding $\mathbf{q}_j^{\mathrm{R}}$ for the phrase \textit{``spoon in contact with the right hand"}.
A challenge here is the absence of explicit ground-truth pairings between detected objects and noun phrases. 
To address this, we adopt a Noise-Contrastive Estimation (NCE) framework~\cite{oord2018representation}, which encourages correct image–narration alignment while implicitly learning object–phrase associations. 
The key idea is that the set of object embeddings from an image $\mathbf{I}_g$ should, as a whole, be more similar to its paired narration $\mathcal{N}_g$ than to any other narration $\mathcal{N}_{g'}$ from the same training batch.


Formally, let $
\{\mathbf{w}_{g1}, \mathbf{w}_{g2}, \dots, \mathbf{w}_{gN}\}$ denote the visual embeddings extracted from image $\mathbf{I}_g$, and let  $
\{\mathbf{\tilde{q}}_{h1}, \dots, \mathbf{\tilde{q}}_{h(2M)}\} 
= \{\mathbf{q}_{h1}^{\mathrm{L}}, \dots, \mathbf{q}_{hM}^{\mathrm{L}}, \mathbf{q}_{h1}^{\mathrm{R}}, \dots, \mathbf{q}_{hM}^{\mathrm{R}}\}$ be the re-indexed textual embeddings corresponding to the left- and right-hand noun phrases from narration $\mathcal{N}_h$.
We then compute a similarity matrix $\mathbf{A}^{gh} \in \mathbb{R}^{N \times 2M}$, where each entry measures how well a given visual embedding matches a given textual embedding: 
\begin{align}
\label{eq:matrix_A}
A_{ij}^{gh} 
&= \mathrm{sim}(\mathbf{w}_{gi}, \mathbf{\tilde{q}}_{hj}) = \frac{\mathbf{w}_{gi} \cdot \mathbf{\tilde{q}}_{hj}^\top}{\|\mathbf{w}_{gi}\| \, \|\mathbf{\tilde{q}}_{hj}\|}.
\end{align}
\noindent
Intuitively, $\mathbf{A}^{gh}$ captures all possible pairwise affinities between detected visual regions in image $\mathbf{I}_g$ and the hand-specific noun phrases from narration $\mathcal{N}_h$, forming the basis for our contrastive alignment objective.
%
Given the assumption that each hand can be in contact with at most one object, we enforce that a hand-specific noun phrase is matched to \textit{only one} visual embedding. 
Operationally, this means that for each textual embedding (each column of $\mathbf{A}^{gh}$), we retain only its strongest match across all visual embeddings. 
Formally, we define the vector $\mathbf{B}$ as follows:
\begin{equation}
\label{eq:NCE_B}
B^{gh}_j = \max_iA_{ij}^{gh}
\end{equation}
Intuitively, this step collapses the object–phrase similarity matrix into a single best-match score per phrase, reflecting the most plausible visual grounding for that phrase under our one-object-per-hand assumption.
%
Next, we aggregate the best-match scores in $\mathbf{B}^{gh}$ into a single global similarity score between image $\mathbf{I}_g$ and narration $\mathcal{N}_h$ by averaging over all $2M$ hand-specific noun phrases in $\mathcal{N}_h$:
\begin{align}
\label{eq:sgh}
S_{gh} 
&= \frac{1}{2M} \sum_{j=1}^{2M} B_{j}^{gh} 
= \frac{1}{2M} \sum_{j=1}^{2M} \max_{i} A_{ij}^{gh}.
\end{align}
Here, $S_{gh}$ reflects the overall alignment between the set of detected objects in $\mathbf{I}_g$ and the hand-specific phrases from $\mathcal{N}_h$, under the one-object-per-hand assumption.
We finally employ the symmetric InfoNCE loss~\cite{oord2018representation,radford2021learning} on $S_{gh}$ to encourage alignment between each image and its paired narration, while pushing apart mismatched pairs in both directions, obtaining the loss contribution $\mathcal{L}_\text{NCE}$.

\subsection{Stage 2: Learning Hand-Object Interactions}
With the vision--language alignment module trained in Stage~1, the second stage (Figure~\ref{fig:wish_arch}{c}) transfers this cross-modal knowledge into a purely visual model capable of identifying in-hand objects at inference time, without relying on narration. 
We achieve this by first generating pseudo-labels in the learned alignment space, which serve as supervisory signals for training two specialized heads designed to directly predict, from visual features alone, the objects being manipulated by the left and right hands, respectively.


The first prediction head, referred to as the \textit{contactness} head, outputs a binary score for each (object $i$, hand $k$) pair, answering the question: 
\textit{``Is object $i$ in physical contact with hand $k$?"}. 
While this head can indicate contact likelihood, it does not enforce the strong assumption that each hand interacts with at most one object. 
To impose this winner-takes-all constraint, we introduce a second prediction head, the \textit{matching} head, which addresses the competitive question: 
\textit{``Among all candidate objects, which one is hand $k$ most likely interacting with?"} 
The two heads are complementary: the matching head assigns each hand to its most likely object, while the contactness head verifies whether that object is indeed in contact, thereby handling the \textit{no-contact} case. 
The next paragraphs detail the pseudo-label generation process and the architectures of these heads.

\paragraph{Pseudo-Label Generation}
We define $\mathbf{A}^b \triangleq \mathbf{A}^{gg}$ as the similarity matrix of matched image-noun pairs.
Each entry $A^b_{ij}$ is the cosine similarity between object $i$ and the $j$-th hand-specific phrase of the same sample.  
For each image–narration pair in the batch, we process $\mathbf{A}^b$ to generate pseudo-labels.
First, for each detected object $i$ and each hand $k \in \{\mathrm{L}, \mathrm{R}\}$, we identify the hand-specific noun phrase with the highest similarity to that object. 
To make the hand index explicit, we reshape $A_{ij}^b$ into $A_{ijk}^b$, where the original index $j \in \{1, \dots, 2M_g\}$ is de-flattened into $j \in \{1, \dots, M_g\}$ for the noun phrase and $k \in \{1, 2\}$ for the hand. 
We then define the condensed score $
\overline{A}^{b}_{ik} = \max_{j} A^{b}_{ijk}$,
which represents the strongest textual evidence that object $i$ is in contact with hand $k$.
We hence derive two sets of pseudo-labels:
\begin{enumerate*}[label=\roman*)]
    \item \textbf{Contact Pseudo-Labels ($\mathbf{L}^b$):}  
    Binary labels indicating whether hand $k$ is in contact with object $i$.  
    We threshold the continuous scores $\overline{A}^{b}_{ik}$ using a dynamic threshold $\rho_\gamma$, defined as the $\gamma$-percentile of all $\overline{A}^{b}_{ik}$ in the batch: $L_{ik}^b = \mathbbm{1}\big(\overline{A}_{ik}^b > \rho_\gamma\big)$.
    \item \textbf{Matching Pseudo-Labels ($\mathbf{H}^b$):}  
    Object indices indicating the most likely match for each hand, under the one-object-per-hand assumption: $
    H^b_{k} = \arg\max_{i} \ \overline{A}^b_{ik}$.
\end{enumerate*}

\paragraph{Contactness Head} 
To produce a binary score for each (object $i$, hand $k$) pair, we use a dedicated adapter $\mathrm{MLP}^C$ projecting object features $\mathbf{w}_{bi}$ and hand features $\mathbf{h}_{bk}$ into a contact-specific space. 
The dot product of these projected features yields the contactness logit:
\begin{equation}
C_{ik}^b = \mathbf{w}^C_{bi} \cdot (\mathbf{h}^C_{bk})^\top, 
\quad 
\begin{cases} 
    \mathbf{w}^C_{bi} = \mathrm{MLP}^C(\mathbf{w}_{bi}) \\ 
    \mathbf{h}^C_{bk} = \mathrm{MLP}^C(\mathbf{h}_{bk}).
\end{cases}
\end{equation}
Applying a sigmoid gives the contact probability 
$Q_{ik}^b = \sigma(C_{ik}^b)$. 
We train this head with the binary contact pseudo-labels $\mathbf{L}^b$ using the focal loss~\cite{lin2017focal} to address the natural class imbalance, obtaining the loss contribution $\mathcal{L}_C$.

\paragraph{Matching Head} 
A second adapter, $\mathrm{MLP}^I$, projects object features $\mathbf{w}_{bi}$ and hand features $\mathbf{h}_{bk}$ into an interaction-specific space. 
The dot product of these projected features yields the matching logits:
\begin{equation}
M_{ik}^b = \mathbf{w}^I_{bi} \cdot (\mathbf{h}^I_{bk})^\top, 
\quad 
\begin{cases} 
    \mathbf{w}^I_{bi} = \mathrm{MLP}^I(\mathbf{w}_{bi}) \\ 
    \mathbf{h}^I_{bk} = \mathrm{MLP}^I(\mathbf{h}_{bk})
\end{cases}.
\end{equation}
To select the single most likely object for each hand, we apply a column-wise softmax over $\mathbf{M}^b$, producing probabilities $
P_{ik}^b = \mathrm{Softmax}(\mathbf{M}^b_{:k})_i$.
We train this head with a cross-entropy loss using the matching pseudo-labels $\mathbf{H}^b$ as targets, obtaining the loss contribution $\mathcal{L}_I$.

\subsection{End-to-End Training and Inference}
The entire \ours model is trained end-to-end by jointly optimizing the objectives from both stages. The total loss is a weighted sum of the three components:
\begin{align}
	\mathcal{L}_{tot} = \lambda_N \mathcal{L}_{\text{NCE}} + \lambda_I \mathcal{L}_I + \lambda_C \mathcal{L}_C,
\end{align}
where $\lambda_N$, $\lambda_I$, and $\lambda_C$ are hyperparameters that balance the contributions of the noun-object alignment, interaction matching, and contactness losses, respectively.

At inference time, we discard Stage 1 and the narrations entirely, using only Stage 2 to make predictions.
Specifically, we use the matching head to find the object most likely in contact with each hand. If the contactness head outputs a score below 0.5, the hand is considered not in contact with any objects. 

\begin{table*}[t]
	\centering
	\small
	\begin{adjustbox}{width=\textwidth}
		\begin{tabularx}{\textwidth}{llc *{4}{>{\centering\arraybackslash}X} *{4}{>{\centering\arraybackslash}X}}
			\toprule
			& & & \multicolumn{4}{c}{\textbf{EPIC-Kitchens}} & \multicolumn{4}{c}{\textbf{Ego4D}} \\
			\cmidrule(lr){4-7} \cmidrule(lr){8-11}
			& & Taxonomy & \textit{E} & \textit{L} & \textit{R} & \textit{B}
			& \textit{E} & \textit{L} & \textit{R} & \textit{B} \\
			\toprule
						
			\rowcolor{lightgray}
			\multicolumn{11}{l}{\textit{F) Fully Supervised}} \\
			\rowcolor{mylightgray}
			F1 & HOS~\cite{VISOR2022} & \checkmark & 50.29 & 37.86 & 37.34 & 19.25 & 55.29 & 34.55 & 39.43 & 30.18 \\
			\midrule

            \rowcolor{lightgray}
			\multicolumn{11}{l}{\textit{O) Oracles with Narrations at Inference Time (EK100 only)}} \\
			\rowcolor{mylightgray}
			O1 & GSAM + IoU-contact & \checkmark                  & 34.33          & 21.78          & 19.29          & 20.06          & -              & -              & -              & -             \\
			\rowcolor{mylightgray}
			O2 & GSAM + LLaVa-contact & \checkmark               & 31.63          & 22.79          & 15.56          & 4.00           & -              & -              & -              & -             \\
			\rowcolor{mylightgray}
			O3 & GSAM + LLaVa-props. &                & 26.70          & 20.64          & 16.38          & 11.90          & -              & -              & -              & -             \\
			\midrule
			
			\rowcolor{mygreen!60}
			\multicolumn{11}{l}{\textit{Z) Zero-shot}} \\
			Z1 & SAM + IoU-contact & & 11.87 & 9.82 & 6.69 & 4.65 & 13.23 & 8.52 & 10.92 & 3.59 \\
			Z2 & GSAM + IoU-contact & \checkmark & 18.72 & 14.98 & 12.06 & 7.81 & 15.81 & 12.42 & 14.79 & 3.62 \\
			Z3 & GSAM + LLaVa-contact & \checkmark & 20.24 & 16.26 & 14.52 & 4.45 & 16.15 & 11.62 & 11.58 & 8.24 \\
			\midrule
			
			\rowcolor{mygreen!60}
			\multicolumn{11}{l}{\textit{D) Distilled from Oracles}} \\
			D1 & HOS~\cite{VISOR2022} \textit{on O1} & & 10.11 & 4.93 & 3.82 & 3.54 & 15.33 & 10.14 & 12.52 & 1.16 \\
			D2 & HOS~\cite{VISOR2022} \textit{on O2} & \checkmark & 15.05 & 16.32 & 4.35 & 1.98 & 16.00 & 12.75 & 5.95 & 0.00 \\
			D3 & HOS~\cite{VISOR2022} \textit{on O3} & \checkmark & 13.01 & 8.91 & 10.84 & 3.05 & \textbf{24.56} & 13.20 & 18.94 & 9.15 \\
			\midrule

            \rowcolor{mygreen!60}
			\multicolumn{11}{l}{\textit{W) Ours}} \\
			W1 & \ours SAM & & 21.31 & 15.73 & 11.91 & 13.17 & 22.55 & 13.78 & 16.51 & \textbf{9.96} \\
			W2 & \ours GSAM & \checkmark & \textbf{27.66} & \textbf{19.49} & \textbf{15.63} & \textbf{13.55} & 23.61 & \textbf{14.23} & \textbf{20.18} & \textbf{9.96} \\
						
			\bottomrule
		\end{tabularx}
	\end{adjustbox}
	\caption{Results on EK100 and Ego4D datasets on the Either (E), Left (L), Right (R), Both (B) mIoU metrics. Oracles cannot be evaluated on Ego4D as narrations are not available on the EgoHOS test set. \textit{Taxonomy} checkmark indicates that the method employs the original object taxonomy from the underlying dataset to work.}
	\label{tab:results}
\end{table*}

\section{Experiments}

\subsection{Compared Baselines}
We compare \ours with different baselines. 
The first baseline is a state-of-the-art Hand-Object Segmentation (HOS)~\cite{VISOR2022} method fine-tuned on fully-supervised annotations (section F in Table~\ref{tab:results}). The second set of baselines (section \textit{Z} in Table~\ref{tab:results}) employs zero-shot vision-language understanding capabilities of state-of-the-art open-vocabulary object detectors like Grounded-SAM (GSAM) \cite{ren2024grounded} and large vision-language models like LLaVa \cite{liu2023visual}. In particular, \textbf{IoU-contact} (Z1, Z2) employs SAM and GSAM using the dataset-specific taxonomy, respectively, to produce object segmentations; then it leverages the IoU scores to determine if the hand is in contact with the object; \textbf{LLaVa-contact (Z3)} is similar to Z2, but employs LLaVa as a visual reasoner to understand which objects are actually interacting with the hand. The third set of baselines (section \textit{D}) employs narration-driven pseudo-labels to train a state-of-the-art HOS model~\cite{VISOR2022}. The pseudo-labels are generated through similar pipelines employed in \textit{Z}, with the difference that they have access to the narrations at inference time. For this reason, we call these pseudo-label generators \textit{oracles} (section \textit{O}). As in the case of F1, oracles are not directly comparable to other NS-iHOS methods due to the extra input or supervision.
O1 and O2, are similar respectively to Z2 and Z3, but employ narrations at inference time instead of the specific dataset taxonomy; 
O3, instead, employs LLaVa with the frame narration injected within the context to produce a suitable taxonomy to feed GSAM, and then leverages the IoU score to determine the interactions.
For fair comparisons, we instantiate \ours,  with both SAM~\cite{kirillov2023segment} and GSAM~\cite{ren2024grounded}.
More details are reported in Appendix.


\begin{figure*}[t]
    \centering
    \includegraphics[width=1\linewidth,clip,trim=0cm 6.4cm 0cm 0cm]{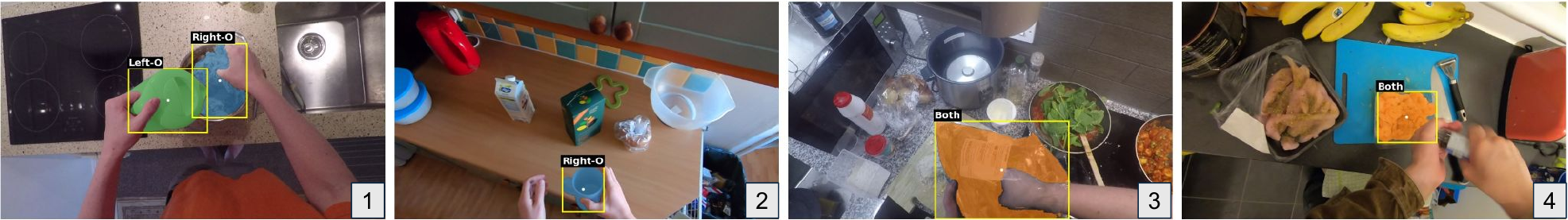}
    \caption{Qualitative results of \ours} 
    \label{fig:qualitative}
\end{figure*}

\subsection{Results}

\paragraph{EPIC-Kitchens} On the left part of Table \ref{tab:results}, we can notice that our method (\ours) achieves the best overall performance, significantly outperforming zero-shot and distillation-based approaches, and even surpassing some oracle baselines that have access to narrations at inference time in the \textit{Both} metric. Compared to the fully supervised HOS~\cite{VISOR2022}, \ours recovers about $50\%$ of the performance despite not relying on manually annotated data during training.
Oracle methods (\textit{O1–O3}) act as upper bounds for  related zero shot methods  by exploiting narrations at inference time, which are typically not available in real-world applications. Their strong performance in some categories highlights the potential of such privileged information, though it limits their practical applicability. Notably, \textit{O1} (GSAM + IoU-contact) achieves the best results among the oracles, indicating that simple spatial heuristics combined with narration-driven segmentation models can be extremely effective. 
GSAM variants outperform SAM variants (both in \ours and in the baselines) across all metrics, highlighting the benefit of using GSAM over the standard SAM. In particular, the taxonomy awareness of GSAM helps generate more semantically relevant proposals, improving alignment with narrated actions.

\paragraph{Ego4D} In the right section of Table \ref{tab:results}, we observe similar trends as in \textit{EPIC-Kitchens}: \ours outperforms zero-shot and distillation-based baselines across most categories, achieving particularly strong results in the \textit{Right-O} and \textit{Both} classes.
As in the previous benchmark, \ours clearly benefits from using GSAM over SAM, highlighting the role of taxonomy-aware proposals. Distilled models show competitive performance, especially when trained on richer pseudo-labels.
In particular, \textit{D3} outperforms our method on the generic \textit{Either} metrics, but still falls short when accurate discrimination between left/right/both is required.
Overall, these results highlight the effectiveness and robustness of \ours, demonstrating strong generalization across scenarios without manual supervision or inference-time narrations—making it a promising solution for NS-iHOS.

\subsection{Ablations and Qualitative Results} 
\paragraph{Role of different components}
In Table \ref{tab:ablations-losses}, we validated the two main components of our architecture by enabling or disabling loss functions for Stage 1 ($\mathcal{L}_\text{NCE}$) and Stage 2 ($\mathcal{L}_I$ and $\mathcal{L}_C$). 
Results highlight that stages can also work in isolation. However, end-to-end training that employs both components is crucial for achieving the best performance.

\paragraph{Pseudo-labels evaluation}
In Table \ref{tab:ablations-pseudo-label-generators}, we measured the quality of the pseudo-labels from the first stage of our architecture. To perform this measurement, we require the assumption that narrations are available at inference time. We reported the results using both SAM and GSAM detectors, comparing the generated pseudo-labels with the final outputs from the second stage. The pseudo-labels are overall weaker than the final results: the second stage learns to distill relevant signals from the narration-guided pseudo-labels, which may contain strong noise. 

\paragraph{Qualitative Results} 
Figure~\ref{fig:qualitative} presents qualitative results. 
Our approach successfully segments objects of varying sizes, identifying both large (image 3) and small ones (image 2), demonstrating robustness to scale variations. \ours also handles complex scenes with multiple objects (image 1) and bimanual interactions (image 3). In the fourth image, \ours mistakenly identifies a non-interacted object, showing that distinguishing between interacted and nearby objects remains challenging. In the Appendix, we also report qualitative results for pseudo-labels.

					

\begin{table}[t]
	\centering
	\small
	\begin{tabular}{ccc cccc}
		\toprule
		$\mathcal{L}_\text{NCE}$ & $\mathcal{L}_\text{C}$ & $\mathcal{L}_\text{I}$ & \textit{E} & \textit{L} & \textit{R} & \textit{B} \\
		\toprule
		\checkmark & & & 16.61 & 11.87 & 10.35 & 5.45 \\
		& \checkmark & \checkmark & 21.85 & 16.42 & 14.14 & 8.59 \\
		\checkmark & \checkmark & & 16.06 & 11.12 & 9.37 & 5.50 \\
		\checkmark & \checkmark & \checkmark & \textbf{27.66} & \textbf{19.49} & \textbf{15.63} & \textbf{13.55} \\
		\bottomrule
	\end{tabular}
	\caption{Effect of the three losses on the EPIC-Kitchens dataset.}
	\label{tab:ablations-losses}
\end{table}

\begin{table}[t]
	\centering
	\small
	\begin{tabular}{l cccc}
		\toprule
		& \textit{E} & \textit{L} & \textit{R} & \textit{B} \\
		\toprule
        \rowcolor{lightgray}
        \multicolumn{5}{l}{\textit{Pseudo-labels from Stage-1 (Narrations at Inference Time)}} \\
		\ours SAM S1 & 14.73 & 11.04 & 8.98 & 10.06 \\
		\ours GSAM S1 & 26.28 & 17.82 & 14.71 & \textbf{13.83} \\
        \midrule
        \rowcolor{mygreen!60}
        \multicolumn{5}{l}{\textit{Final output from Stage-2}} \\
        \ours SAM & 21.31 & 15.73 & 11.91 & 13.17 \\
		\ours GSAM & \textbf{27.66} & \textbf{19.49} & \textbf{15.63} & 13.55 \\
		\bottomrule
	\end{tabular}
	\caption{Performance of pseudo-labels (obtained by employing narrations at inference time) on the EPIC-Kitchens dataset.}
	\label{tab:ablations-pseudo-label-generators}
\end{table}


\section{Conclusions}

We introduced NS-iHOS, a novel task for in-hand object segmentation from egocentric images supervised only by natural language narrations. We proposed a two-stage framework that uses CLIP-based alignment to generate pseudo-labels and learn hand-object associations without the need for narrations at test time. Experiments on EPIC-Kitchens and Ego4D show that our model outperforms baselines and recovers over 50\% of the performance of fully supervised methods. Our work establishes a new benchmark for scalable, narration-driven understanding, with future plans to tackle egocentric \textit{videos} and larger datasets.

\section*{Acknowledgments}
This work was funded by Spoke 8, Tuscany Health Ecosystem (THE) Project (CUP B83C22003930001), funded by the NRRP within the NextGeneration Europe (NGEU) Program; SUN -- Social and hUman ceNtered XR (EC, Horizon Europe No. 101092612); Future Artificial Intelligence Research (FAIR) – PNRR MUR Cod. PE0000013 - CUP: E63C22001940006. We thank CINECA under the ISCRA initiative, for HPC resources and support (GEPPETHO project).

\bibliographystyle{elsarticle-num} 
\bibliography{bibliography,bibliography_VL}

\appendix
\section{\ours{} details}

\subsection{Implementation Details}
\label{sec:implementation-details}
We train the few parameters of \ours for 15 epochs employing the Adam optimizer with a learning rate of $4\cdot 10^{-6}$. We empirically set $\lambda_N=0.2$, $\lambda_I=0.1$, and $\lambda_C=1.0$ -- following the criteria of equally balancing the magnitudes of the three losses. We also employ the percentile threshold $\gamma = 0.3$ and the focal loss modulating factor $\theta = 2$. In order to decrease noise in the pseudo-label generation, we also mask out from the pseudo-labels $\mathbf{L}$ and $\mathbf{H}$ all the entries for which $\text{IoU}(\mathbf{m}_i, \mathbf{m}_k^\text{hand}) = 0$ -- i.e., all the entries relative to objects whose mask $\mathbf{m}_i$ do not intersect with the hand mask $\mathbf{m}^\text{hand}_k$. As object detectors, we employ SAM~\cite{kirillov2023segment} and GSAM~\cite{ren2024grounded}, prompted with the dataset-specific taxonomy to adhere to the baselines, in both cases using \cite{leonardi2025synthetic} as the hand detector. 
The configuration is the same for the two datasets.

\subsection{Visual and Textual Backbones}
\paragraph{Visual Backbone}
The visual backbone is shared across both stages of \ours.
Given an input image $\mathbf{I} \in \mathbb{R}^{W\times H\times 3}$, our first goal is to localize all candidate objects and hands that may participate in an interaction.
To this end, we apply a pre-trained, class-agnostic object segmenter $\mathcal{D}$, producing a set of $N$ binary masks  $\{\mathbf{m}_1, \mathbf{m}_2, \dots, \mathbf{m}_N\}$,
each highlighting a distinct object or hand region.
We embed these regions into a semantically rich feature space using the vision-language model CLIP~\cite{radford2021learning}.
Specifically, for each mask $\mathbf{m}_i$, we compute a visual embedding $
\mathbf{v}_i \in \mathbb{R}^{D_v}, \quad \mathbf{v}_i = \mathrm{CLIP}^v(\mathbf{I}, \mathbf{m}_i)$,
where $\mathrm{CLIP}^v$ is a mask-guided variant of the CLIP image encoder~\cite{shen2024learning}, which applies the mask $\mathbf{m}_i$ to the last attention layer, ensuring that the resulting CLS token captures object-specific features while still being contextualized by the entire scene.



We apply the same procedure to left and right hands detected through an off-the-shelf hand detector (see Section \ref{supp:hand_detector}), which are then encoded by the mask-guided CLIP encoder to yield hand embeddings $\mathbf{h}_{\mathrm{L}}$ and $\mathbf{h}_{\mathrm{R}}$.
During training, we update only the MLP adapters, keeping the CLIP backbone and object/hand detectors frozen.

\paragraph{Textual Backbone} 
%
To connect visual representations with the narration $\mathcal{N}$, we first extract its key semantic entities. 
Using standard NLP tools (see Section \ref{supp:noun-extraction}), we parse $\mathcal{N}$ to obtain a set of $M$ noun phrases $
\{\mathcal{P}_1, \mathcal{P}_2, \dots, \mathcal{P}_M\}$,
each corresponding to a distinct object or entity mentioned in the narration.
For example, given the narration \textit{``I pick the scraper from the bucket of water"}, the extracted phrases would include \textit{``scraper"} and \textit{``bucket of water"}.


However, the extracted noun phrases are generic and do not indicate which hand, if any, is interacting with the referenced object. 
A key responsibility of the first stage is therefore to derive potential object–hand associations, compensating for this missing information in the narration.
To this aim, we enrich each noun phrase $\mathcal{P}_j$ using two hand-specific prompt templates: $\mathcal{F}^L$ (\textit{``[noun] in contact with the left hand"}) and $\mathcal{F}^R$ (\textit{``[noun] in contact with the right hand"}). For example, from $\mathcal{P}_j=\textit{``spoon''}$, we generate two distinct phrases: $\mathcal{P}_j^L$=\textit{``spoon in contact with the left hand''} and $\mathcal{P}_j^R$=\textit{``spoon in contact with the right hand''}.
This augmentation step explicitly encodes possible hand–object relationships in language form, enabling the model to later align visual regions with the correct interaction type.

These augmented phrases are then encoded using the pre-trained CLIP text encoder $\text{CLIP}^t$ to get embeddings $\mathbf{p}_j^L$ and $\mathbf{p}_j^R$. 

\subsection{Hand Detector}
\label{supp:hand_detector}
We adopt a hand detector based on the architecture introduced in~\cite{leonardi2025synthetic}, designed to operate in egocentric settings. The model is trained using a combination of synthetic labeled hand data and real-world domain-specific unlabeled data, following a semi-supervised training scheme. We use this detector to extract hand predictions both for our method and for the various baselines, ensuring a consistent and fair comparison across all evaluated approaches. To ensure consistent hand-side prediction, we apply a simple heuristic to assign a \textit{left} or \textit{right} label to each detected hand. The image is vertically divided into two halves: hands detected in the left half are designated as \textit{left}, while those in the right half are identified as \textit{right}. In cases where two hands are detected, the leftmost is assigned the \textit{left} side, and the other is assigned the \textit{right} side. Despite the simplicity of this heuristic, we measured on the VISOR benchmark that the hand-side mAP is 85.5, meaning that on egocentric frames, this is enough to distinguish the left/right sides with high accuracy. 

\subsection{Nouns Extraction from Narrations}
\label{supp:noun-extraction}
EPIC-Kitchens dataset already comes with precomputed nouns extracted from the narration employing off-the-shelf part-of-speech (POS) taggers. These nouns are already available in the official EPIC-Kitchens metadata. Concerning the narrations from Ego4D, we pre-process the narrations to eliminate the user identifier (\textit{\#C C} at the beginning of the text) and we then perform POS tagging employing the \texttt{nltk} Python framework.

\section{Zero-shot Baselines and Oracles}
\label{sec:baselines-oracles}
In this section, we better clarify the construction of our zero-shot baselines (Z1, Z2, Z3) together with the oracles (O1, O2, O3), both making use of a careful orchestration between SOTA open-vocabulary detection and segmentation models (SAM \cite{kirillov2023segment} and GSAM \cite{ren2024grounded}) and large vision-language models (LLaVa \cite{liu2023visual}) employed as a visual reasoner.

\subsection{Baselines}

\paragraph{Fully Supervised Baseline}
We train a state-of-the-art HOS~\cite{VISOR2022} method using fully supervised labels.
For EPIC-KITCHENS, we use the labels available in the training set.
Since we do not have fully supervised labels in the training set of the Ego4D subset, we rely on the fully supervised training set of EgoHOS~\cite{EgoHos_jianbo_eccv22} to train this baseline.

\paragraph{Z1 - \textbf{SAM + IoU-contact}.}
This baseline uses the Segment Anything Model (SAM) to generate instance segmentations of the image without any object category information. Since SAM lacks semantic understanding, it is most likely to produce larger noise, but, unlike open-vocabulary detectors, it avoids the dependency on specific taxonomies. We determine which segmented objects are in contact with the left or right hand by computing the Intersection-over-Union (IoU) between each segment and the hand mask. For each hand, the object with the highest IoU is selected as the potential interacting object.

\paragraph{Z2 - \textbf{GSAM + IoU-contact} (EK-100 Taxonomy).}
This method builds on the previous baseline by replacing SAM with Grounded-Segment-Anything (GSAM), which extends SAM with open-vocabulary object grounding. We supply GSAM with a predefined taxonomy derived from the Ego4D and EPIC Kitchens dataset to guide object detection and segmentation. As in Z1, the interaction decision is made by computing IoU between the hand and the segmented objects. The object with the highest IoU is selected for each hand.

\paragraph{Z3 - \textbf{GSAM + LLaVa-contact} (EK-100 Taxonomy).}
In this baseline, we combine GSAM with the large vision-language model LLaVa to determine hand-object interactions through visual reasoning. First, GSAM is prompted with the EK-100 taxonomy to extract object proposals. These detected objects are then listed and injected into a prompt to LLaVa, which is tasked with assessing, for each hand separately, which object from the list it is most likely in contact with. The prompt used employs a simple chain-of-thoughts which makes LLaVa aware of the context before asking to produce the final output:
\begin{figure*}[t]
\centering
\begin{tcolorbox}[colback=gray!5!white, colframe=black!75, title=Prompt used in Z3 (GSAM + LLaVa-contact), fonttitle=\bfseries]
\footnotesize
\textbf{Task Description:} \\
The frame shows a first-person view from a camera mounted on the user's head. I would like to understand with which objects each of the two hands, if visible, are in contact.

\vspace{0.5em}
\textbf{Left Hand Query:} \\
The left hand may be in contact with these objects: \texttt{gsam\_dets.all().intersect(gsam\_dets.left\_hand())}.
With which of the listed objects is the left hand most likely in contact, looking at the image? And with which confidence (from 0.0 to 1.0)? If the left hand is not visible or not touching anything, please state so. Be concise and brief.

\vspace{0.5em}
\textbf{Right Hand Query:} \\
Now focus on the right hand. The right hand may be in contact with these objects: \texttt{gsam\_dets.all().intersect(gsam\_dets.right\_hand())}. With which of the listed objects is the right hand most likely in contact, looking at the image? And with which confidence (from 0.0 to 1.0)? If the right hand is not visible or not touching anything, please state so. Be concise and brief.

\vspace{0.5em}
\textbf{Output Format:} \\
Formalize your findings in JSON format. If the hand is not touching anything, put \texttt{None} to the object field and \texttt{0.0} to \texttt{contact\_confidence}. The variable \texttt{contact\_confidence} should range between 0.0 (when there is surely no contact) and 1.0 (when there is surely contact). The field \texttt{interacting\_object} should contain one of the listed objects.\\[0.3em]

\textit{Example output:}
\begin{verbatim}
{
  "left_hand": {
    "interacting_object": "cup",
    "contact_state": "in-contact",
    "contact_confidence": 0.8
  },
  "right_hand": {
    "interacting_object": null,
    "contact_state": "no-contact",
    "contact_confidence": 0.1
  }
}
\end{verbatim}
\end{tcolorbox}
\end{figure*}

\subsection{Oracles}
To define upper bounds for weakly supervised learning pipelines, we construct oracles that are granted access to narrations at inference time. These oracles use similar underlying models to the zero-shot baselines but incorporate noun phrases or full narration content to guide object proposal and reasoning, which, in the NS-iHOS task, we suppose to be unavailable at prediction time. These oracles are not valid baselines, as they make use of information not available during actual inference, but provide insight into the upper-bound performance of narration-guided systems.

\paragraph{O1 - \textbf{GSAM + IoU-contact} (Only narrations)}
\textbf{O1} uses GSAM guided exclusively by nouns extracted from the narration instead of the specific taxonomy to produce object proposals and employs IoU-contact as in Z2.

\paragraph{O2 - \textbf{GSAM + LLaVa-contact} (Only narrations)}
\textbf{O2} also uses narration-derived object proposals from GSAM, but interaction reasoning is delegated to LLaVa as in Z3.

\paragraph{O3 - \textbf{GSAM + LLaVa-proposals-and-narrations}}
\textbf{O3} prompts LLaVa with the full narration to dynamically construct a scene-specific object vocabulary. This vocabulary is then passed to GSAM, whose segmentations are interpreted through IoU-contact logic. In this case, LLaVa is used for a different purpose with respect to Z3, and the employed prompt is reported below.

\begin{tcolorbox}[colback=gray!5!white, colframe=black!75, title=Prompt used to extract object list in O3, fonttitle=\bfseries]
\footnotesize
\textbf{Scene Understanding:} \\
What is happening in this egocentric scene? Be concise and brief, but give all the details about objects and the actions happening. Consider this ground truth description: \texttt{narration}. Place particular attention to the possibly tiny objects already grabbed by the hands.

\vspace{0.5em}
\textbf{Object Selection Criteria:} \\
List at most 6 objects, giving precedence to the ones in proximity to the hands. Do not consider hands, arms, or any human body parts.

\vspace{0.5em}
\textbf{Output Format:} \\
Print the object names as a one-line string of labels separated by commas.
\end{tcolorbox}

\section{Pseudo-label Qualitative Results}
\begin{figure*}[t]
    \centering
    \includegraphics[width=1\linewidth]{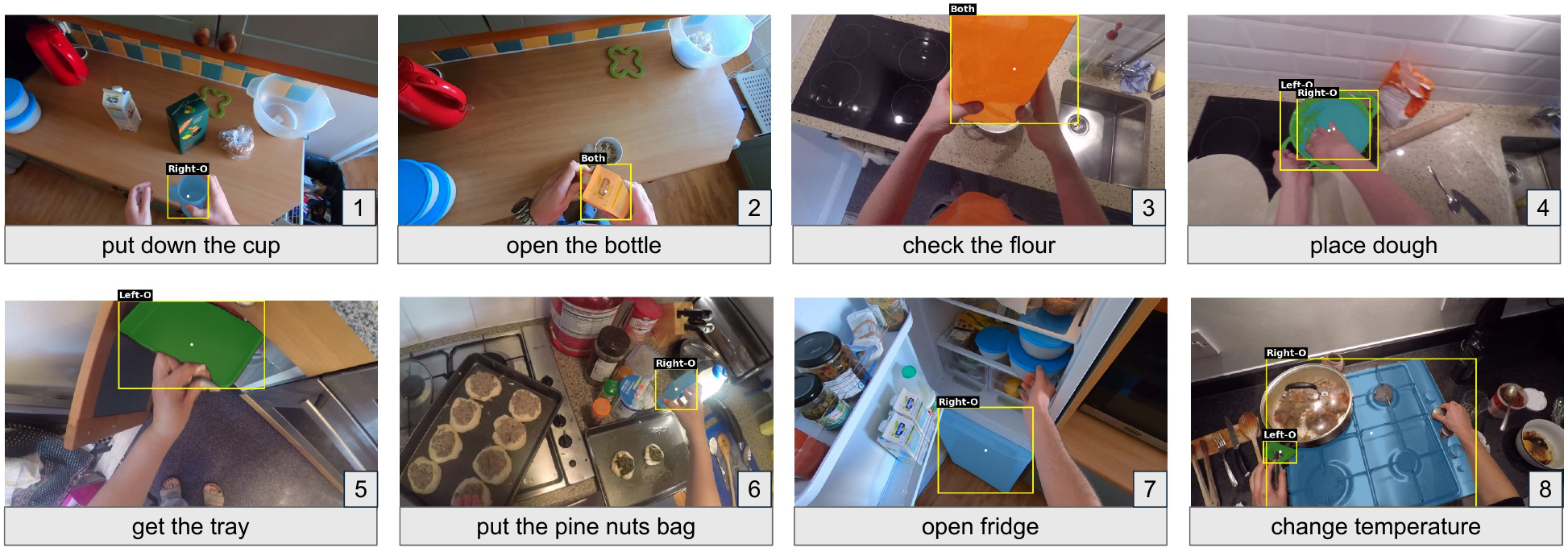}
    \caption{Qualitative results of the pseudo-labels from Stage 2 of \ours on EPIC‑Kitchens.}
    \label{fig:qualitative-pseudo}
\end{figure*}

In Figure \ref{fig:qualitative-pseudo}, we show some qualitative results from Stage 2, which is the stage of \ours dedicated to the production of pseudo-labels. We report narrations underneath every figure, as these pseudo-labels are obtained by aligning hand-specific textual representations with the objects in the image, and therefore require narrations to be produced.
As we can notice, the model can effectively ground the narration into the image, also understanding the interaction with the hands. Notably, \ours can produce meaningful pseudo-labels even in cases where the narration does not directly reference a concrete object in the scene (like in example 8, where CLIP is anyway able to deduce that \textit{temperature} refers to the cooker). In example 7, we have a failure case caused by the narration not always being perfectly aligned with the video frame. In that moment, the person had already opened the fridge and was instead interacting with its contents.

\end{document}